# Action Recognition with Dynamic Image Networks

Hakan Bilen, Basura Fernando, Efstratios Gavves, and Andrea Vedaldi

**Abstract**—We introduce the concept of *dynamic image*, a novel compact representation of videos useful for video analysis, particularly in combination with convolutional neural networks (CNNs). A dynamic image encodes temporal data such as RGB or optical flow videos by using the concept of 'rank pooling'. The idea is to learn a ranking machine that captures the temporal evolution of the data and to use the parameters of the latter as a representation. When a linear ranking machine is used, the resulting representation is in the form of an image, which we call dynamic because it summarizes the video dynamics in addition of appearance. This is a powerful idea because it allows to convert any video to an image so that existing CNN models pre-trained for the analysis of still images can be immediately extended to videos. We also present an efficient and effective approximate rank pooling operator, accelerating standard rank pooling algorithms by orders of magnitude, and formulate that as a CNN layer. This new layer allows generalizing dynamic images to dynamic feature maps. We demonstrate the power of the new representations on standard benchmarks in action recognition achieving state-of-the-art performance.

✦

## 1 INTRODUCTION

Videos account for a large majority of the visual data in existence, surpassing by a wide margin still images. Therefore understanding the content of videos accurately and on a large scale is of paramount importance. The advent of modern learnable representations such as deep convolutional neural networks (CNNs) has improved dramatically the performance of algorithms in many image understanding tasks. Since videos are composed of a sequence of still images, some of these improvements have been shown to transfer to videos directly. However, it remains unclear *how videos can be optimally represented*. For example, a video can be represented as a sequence of still images, as a subspace of images or image features, as the parameters of a generative model of the video, or as the output of a neural network or even of an handcrafted encoder.

Early works [12], [30], [68] represented videos as (the parameter of) models generating them. Doretto *et al.* [12] introduced the concept of *dynamic textures*, reconstructing pixel intensities as the output of an auto-regressive linear dynamical system. Wang *et al.* [68] used instead the moments of a mixture of Gaussians generating temporally local, flow-based appearance variations in the video.

More recent approaches [15], [21], [44], [57] focus on the problem of understanding the content of videos, which does not necessarily requires to model their dynamics. They do so by

- Hakan Bilen is with the School of Informatics, University of Edinburgh. E-mail: hbilen@inf.ed.ac.uk
- Basura Fernando is with the ACRV, Research School of Engineering, The Australian National University, ACT 2601, Australia. E-mail: basura.fernando@anu.edu.au
- Efstratios Gavves is with the QUVA Lab, University of Amsterdam. E-mail: egavves@uva.nl
- Andrea Vedaldi is with the VGG, University of Oxford. E-mail: vedaldi@robots.ox.ac.uk

1. From left to right and top to bottom: "blowing hair dry", "band marching", "balancing on beam", "golf swing", "fencing", "playing the cello", "horse racing", "doing push-ups", "drumming".

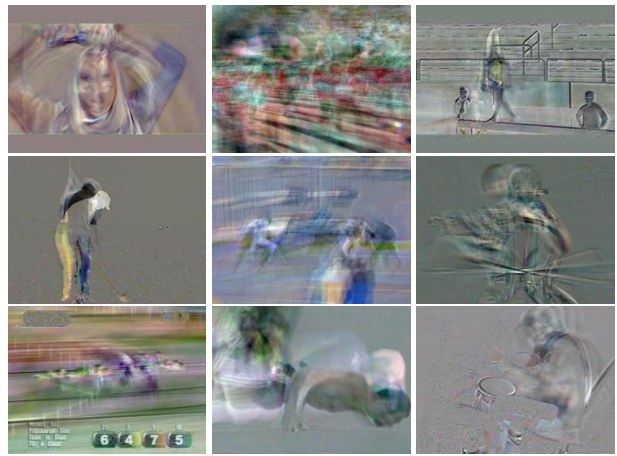

Fig. 1: Examples of *dynamic images* summarizing short video sequences as still images. They provide a simple, powerful, and efficient representation of videos for action recognition. *Can you guess what actions are visualized?*[1]

treating videos as stack of frames and then learning discriminative models that distill the information needed to solve specific problems such as action recognition. The majority of these methods rely on convolutional or recurrent neural networks and learn spatio-temporal filters that maximize the recognition capability of the overall system in an end-to-end manner. This allows these approaches to achieve the highest accuracy in action recognition, as their primary purpose is to model the action classes and not the motion itself.

In this paper, we propose a new representation of videos that, as in the first examples, encodes the data in a general and content-agnostic manner, resulting in a long-term, robust motion representation applicable not only to action recognition, but to other video analysis tasks as well [39], [64]. This new representation distills the motion information contained in all the frames of a video into a single image, which we call the *dynamic image*. We show that the dynamic image is a powerful, efficient, and yet simple representation of videos, particularly useful in the context of deep learning.

A popular method to represent time series is to apply a *temporal pooling operator* to the features extracted at individual time instants. For videos, temporal pooling has been done by using temporal templates [3], ranking functions for video frames [17] and sub-videos [23], as well as more traditional pooling operators [54]. CNN add another dimension to this research, as one has to decide where pooling should take place. A CNN such as AlexNet [32] contains in fact a whole hierarchy of image representations, one for each layer in the network. One could pool the output of the deep fully-connected layers of the network, but this would prevent the CNN from analyzing the video dynamics. Alternatively, temporal pooling could be applied to some intermediate network layer. In this case, the lower layers would still observe single frames, but the upper layers could reason about the overall video dynamics.

The dynamic image (section 3) takes this idea to its logical extreme and captures the video dynamics *directly at the level of the image pixels*, by applying a pooling operator before any of the CNN layers are evaluated. The dynamic image is a single RGB image, equivalent to a still, that captures the gist of the dynamics and appearance of a whole video sequence or subsequence (fig. 1). The dynamic image is obtained as the parameter of ranking machine learned to sort the frames of the video temporally, a method proposed by [17], [18]; the key difference from this prior work is that the ranking machine is computed directly at the level of the image pixels as well as any intermediate level of a CNN feature extractor.

This idea has four keys advantages. First, the dynamic image can be processed by any of the many CNN architecture for still images while still being able to reason about the long-term dynamics in a video. Second, the dynamic image is very efficient: extracting it is simple and quick, and reducing the analysis of videos to the analysis of a single RGB images significantly accelerates recognition. Third, the representation is very compact, as a whole video is summarized by an amount of data equivalent to a single frame. Compressing videos in this manner is very useful for large scale indexing. Fourth, dynamic images can be generalized to different kinds of sequences and to different modalities, as we demonstrate by applying it to optical flow frames (section 5.9).

Our second contribution in this paper is to provide a fast approximation to learning the ranking machine which is needed to extract dynamic images. This approximation, which we call *approximate rank pooling* (ARP), amounts to a simple weighted summation of the video frames where the weights are fixed for all videos of the same length and can therefore be pre-computed. This makes ARP an extremely efficient in practice.

ARP defines a map from sequences of $N$-video frames $(I_{\sigma(1)}, \ldots, I_{\sigma(N)})$ presented in an order $\sigma$ to a single dynamic image $I_{\text{dyn}}$. Unlike other commonly used temporal pooling operators like max- or average-pooling that are orderless, and therefore, time/sequence invariant, ARP is sensitive to the permutation order $\sigma$. To the best of our knowledge, we are the first to propose a temporal pooling layer for neural network architectures that is *sensitive to the order* of the samples within a video sequence. We show that ARP can be seamlessly integrated into the end-to-end training of CNNs for video data. We also show that ARP can be applied to the intermediate layers of a CNN too, which can be used to obtain a multi-scale representation of videos.

As a third contribution, we demonstrate the power of the dynamic image and of ARP by applying them to the recognition of human actions in video sequences. Recent works such as [16], [17], [18], [23], [54] pointed out that long term dynamics and temporal patterns are very important cues for the recognition of actions. However, representing complex long term dynamics is challenging, particularly if one seeks compact representations that can be processed efficiently. We do so by introducing a hybrid model (section 4) that makes use of both static and dynamic images and pools information from RGB and optical flow frames from short and long video subsequences. This results in a novel *four stream architecture* that can efficiently and accurately recognize actions in videos, obtaining state-of-the-art performance in standard benchmarks (section 5).

This paper is an extended version of our prior conference publication [2]. The new contributions are:

- a more extensive overview and comparison of the related literature,
- a more detailed formulation of the proposed pooling operations in section 3.2,
- a novel four-stream architecture, adding two new dynamic image streams using optical flow input to the standard two-stream architecture of [57] in section 4.4,
- the use of more powerful deep networks ResNeXt-50 and ResNeXt-101 [75] which result in significantly improved baseline action classification performance section 5,
- a thorough evaluation in section 5 of the proposed ARP, when applied to intermediate layers of a CNN instead of RGB pixels, obtaining state-of-the-art action classification accuracies in popular benchmarks ,
- an alternative temporal pooling strategy, called *parametric pooling*, whose parameters can be automatically learned together with the other network parameters in section 5.7,
- a detailed analysis of various design choices such as temporal window length, sampling rate, temporal pooling strategies in section 5,
- an extended qualitative and quantitative comparison to the previous work (Motion History Images [3]) in section 5.

The rest of the paper is organized as follows: Section 2 provides an extensive overview of related work in video modeling and action recognition. Section 3 formulates the dynamic image and approximate rank pooling. Section 4 proposes different deep neural network architectures using dynamic images and explains how the proposed pooling operators can be integrated into standard CNN models. Section 5 provides a rigorous analysis of the design choices and evaluates the performance of the proposed models in standard human action recognition benchmarks. Section 6 summarizes our findings and discusses future directions.

## 2 RELATED WORK

**Videos as stack of still images:** Existing video representations can be grouped into two categories. The first one, which comprises the majority of the literature on video processing, action and event classification, be it with shallow [17], [18], [43], [68] or deep representations [44], [57], considers videos either as a stream of still images [44] or as a short and smooth transition between similar frames [57]. [44] show that treating videos as bag of static frames performs reasonably well for recognition, as the context of an action typically correlates with the action itself (*e.g.*, "playing basketball" usually takes place in a basketball court).

**Videos as spatio-temporal volumes:** The second category considers videos as 3D dimensional volumes instead of collections of 2D images. Before deep learning became popular, several authors [50], [52], [56] proposed to learn spatio-temporal templates from such spatio-temporal volumes. More recent works [27], [65] extend spatial CNNs to a third, temporal dimension [27] substituting 2D filters with 3D ones. Tran *et al.* [65] show that 3D convolutional networks perform well in the presence of large amount of annotated videos. While the extension brings a more natural representation of videos, it leads to a significant increase in the number of parameters to learn and thus requires more training data. Furthermore, such representations do not account for the fact that the third dimension, time, is not homogeneous with the first two, space.

Simonyan *et al.* [57] show an alternative way of exploiting spatio-temporal information in videos by training a deep neural networks on pre-computed optical flow rather than raw RGB frames and report significant improvements over previous state-of-the-art. Similarly, [21] uses action tubes to to fit a double stream appearance- and motion-based neural network that captures the movement of an actor.

**Short and long-term dynamics:** While the aforementioned methods successfully capture the local changes within a small time window, they cannot capture longer-term motion patterns associated with certain actions. An alternative solution is to consider a second family of architectures based on recurrent neural networks (RNNs) [11], [61]. RNNs typically consider memory cells [24], which are sensitive to both short as well as longer term patterns. RNNs parse the video frames sequentially and encode the frame-level information in their memory. In [11] LSTMs are used together with convolutional neural network activations to either output an action label or a video description. In [61] an autoencoder-like LSTM architecture is proposed such that either the current frame or the next frame is accurately reconstructed. Finally, the authors of [77] propose an LSTM with a temporal attention model for densely labeling video frames.

Many of the ideas in video CNNs originated in earlier architectures that used hand-crafted features. For example, the authors of [36], [67], [68] have shown that local motion patterns in short frame sequences can capture very well the short temporal structures in actions. The rank pooling idea, on which our dynamic images are based, was proposed in [17], [18] using hand-crafted representation of the frames.

**Multi-stream networks:** Our static/dynamic CNN uses a multi-stream architecture. Multiple streams have been used in a variety of different contexts. Examples include Siamese architectures for learning metrics for face identification [6], for unsupervised training of CNNs [10] or, for training externally a visual object tracker to track by searching instances [64]. Simonyan *et al.* [57] use two streams to encode respectively static frames and optical flow frames in action recognition. Recently, Feichtenhofer *et al.* [15] show that fusing two streams via a 3D convolution further improves the classification performance. The authors of [41] propose a dual loss neural network, where coarse and fine outputs are jointly optimized. A difference of our model compared to these is that we branch off two streams at arbitrary location in the network, either at the input, at the level of the convolutional layers, or at the level of the fully-connected layers.

**Motion information:** Motion is a rich source of information for recognizing human actions. Kinematic feature design is heavily studied in the context of human action recognition. In this regard, techniques such as motion summary methods [3], optical flow [1], [13], [29], [55], and dynamic textures [30] are used to capture motion.

Our work is also related to early work on motion summary techniques such as motion energy image (MEI) and motion history image (MHI) [3]. Given an image sequence, the binary MEIs highlight regions in the image where any form of motion was present. To construct MEIs, the summation of the square of consecutive image differences is used as a robust spatial motion-distribution signal. To encode the motion of an image sequence, the motion history images (MHI) are used. In an MHI, pixel intensity is a function of the motion history at that location, where brighter values correspond to more recent motion. As a matter of fact, we compare the proposed method to MHI quantitatively and qualitatively in section 5.

Optical-flow based methods estimate the optical-flow between successive frames and then summarize the motion using principle components [1], [55]. In some instances the optical flow is computed on sub-volumes of the whole video using integral videos [29], or surrounding the central motion [13]. However, normally, the optical-flow, provides only the local dynamics and aggregation of local motion is performed using simple summarization methods.

**Spatio-temporal dynamics:** Dynamic texture [12] uses auto-regressive moving average process which estimates the parameters of the model using sequence data. Dynamic textures methods evolved from techniques originally designed for recognizing textures in 2D images [12], where they were extended to time-varying "dynamic textures" [30] for sequence recognition tasks. The Local Binary Patterns (LBP) [45], for example, use short binary strings to encode the micro-texture centered around each pixel. A whole 2D image is represented by the frequencies of these binary strings. In [30], [81] the LBP descriptor was extended to 3D video data and successfully applied to facial expression recognition tasks. Subspace-based methods are used in [37]. These methods captured some time-varying information for sequence classification tasks.

Even though these techniques [1], [3], [30], [55] provides a solution to capture motion of video sequences, none of them use a *learning strategy* based on optimization to summarize the motion dynamics of video sequence as our method. Moreover, we are the first to use a motion summary images to train CNNs for human action recognition. Our motion summary concept is based on rank pooling and can be applied at different levels of CNN architecture.

**Learning to rank videos:** More recently the rank pooling [18] method is extended in [16], [20] to increase the capacity of rank pooling using a hierarchical approach. In [19], an end-to-end video representation learning method is proposed using CNNs and rank-pooling. Our method is also based on rank pooling [18], however, compared [18] we learn the video representations end-to-end while being more efficient than [19]. The end-to-end video classification method [19] relies on computing the exact gradient of the rank pooling operator where as we argue that it is a good compromise to approximate the gradient of the rank pooling function considering exact method of [19] has to rely on bi-level optimization [22]. In this paper, we only take the first gradient step of the rank pooling operator which allows us to obtain a reasonable solution to the initial optimization problem. To the best of our knowledge such effective optimization trick has not been tried before in the context of CNN-based learning.

The impact of objects in action recognition is studied in [25]. Fisher vector [49] and VLAD descriptor based action recognition has shown promising results along with hand-crafted features [35], [38], [46]. Attributes [40], action-parts [51], [72], hierarchy [34], [59], [73], trajectory pooled deep features [69], human pose [5], [76] and the context [42] also have been used for action recognition. Overall, the good practices in action recognition is described in [74].

## 3 DYNAMIC IMAGES

In this section we introduce the concept of *dynamic image*, which is a standard RGB image that summarizes the appearance and dynamics of a whole video sequence (section 3.1). Then, we we propose a fast approximation to accelerate the computation of dynamic images (section 3.2).

### 3.1 Constructing dynamic images

While CNNs can learn automatically powerful data representations, they can only operate within the confines of a specific hand-crafted architecture. In designing a CNN for video data, in particular, it is necessary to think of how the video information should be presented to the CNN. As discussed in section 2, standard solutions include encoding sub-videos of a fixed duration as multi-dimensional arrays or using recurrent architectures. Here we propose an alternative and more efficient approach in which the video content is summarized by a single still image. This image can then be processed by a standard CNN architecture such as CaffeNet [28] or ResNeXt [75].

Summarizing the video content in a single still image may seem difficult. In particular, it is not clear how image pixels, which already contain appearance information in the video frames, could be overloaded to reflect dynamic information as well, and in particular the long-term dynamics that are important in action recognition.

We show here that the construction of Fernando *et al.* [17], [18] can be used to obtain exactly such an image. The idea of their work is to represent a video as a *ranking function* for its frames $I_1, \ldots, I_T$. In more detail, let $\psi(I_t) \in \mathbb{R}^d$ be a representation or feature vector extracted from each individual frame $I_t$ in the video. Let $V_t = \frac{1}{t} \sum_{\tau=1}^{t} \psi(I_\tau)$ be time average of these features up to time $t$. The ranking function associates to each time $t$ a score $S(t|\mathbf{d}) = \langle \mathbf{d}, V_t \rangle$, where $\mathbf{d} \in \mathbb{R}^d$ is a vector of parameters. The function parameters $\mathbf{d}$ are learned so that the scores reflect the rank of the frames in the video. Therefore, later times are associated with larger scores, *i.e.* $\forall \{q, t\}$ s.t. $q \succ t \implies S(q|\mathbf{d}) > S(t|\mathbf{d})$. Learning $\mathbf{d}$ is posed as a convex optimization problem using the RankSVM [58] formulation:

$$\mathbf{d}^* = \rho(I_1, \ldots, I_T; \psi) = \operatorname*{argmin}_{\mathbf{d}} E(\mathbf{d}),$$
$$E(\mathbf{d}) = \frac{\lambda}{2}\|\mathbf{d}\|^2 + \qquad (1)$$
$$\frac{2}{T(T-1)} \times \sum_{q>t} \max\{0, 1 - S(q|\mathbf{d}) + S(t|\mathbf{d})\}.$$

The first term in this objective function is the usual quadratic regularizer used in SVMs. The second term is a hinge-loss soft-counting how many pairs $q \succ t$ are *incorrectly* ranked by the scoring function. Note in particular that a pair is considered correctly ranked only if scores are separated by at least a unit margin, i.e. $S(q|\mathbf{d}) > S(t|\mathbf{d}) + 1$.

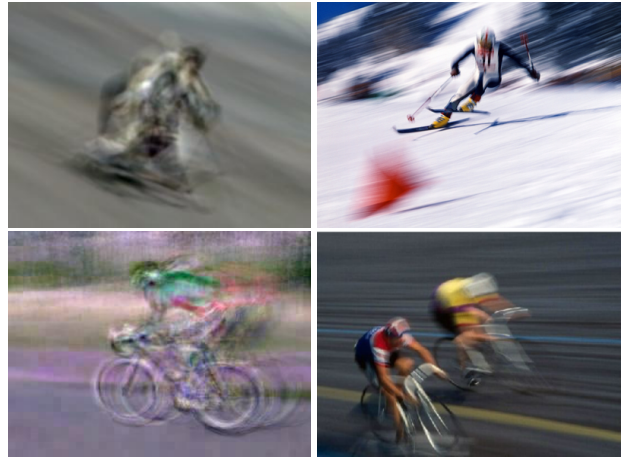

Fig. 2: Left column: dynamic images. Right column: motion blur. Although fundamentally different both methodologically, as well as in terms of applications, they both seem to capture time in a similar manner.

The optimizer to eq. (1) is written as a function $\rho(I_1, \ldots, I_T; \psi)$ that maps a sequence of $T$ video frames to a single vector $\mathbf{d}^*$. Since this vector contains enough information to rank all the frames in the video, it aggregates information from all of them and can be used as a video descriptor. The process of constructing $\mathbf{d}^*$ from a sequence of video frames is known as *rank pooling* [18].

In [17] the map $\psi(\cdot)$ used in this construction is set to be the Fisher Vector coding of a number of local features (histogram of gradients (HOG) [7], histogram of optical flow (HOF) [36], motion boundary histograms (MBH) [8], improved dense trajectories (IDT) [68]) extracted from individual video frames. Here, we propose to apply rank pooling *directly to the RGB image pixels* instead. While this idea is simple, in the next several sections we will show that it has remarkable advantages.

The function $\psi(I_t)$ is now an operator that stacks the RGB components of each pixel in image $I_t$ on a large vector. Alternatively, $\psi(I_t)$ may incorporate a simple component-wise non-linearity, such as the square root function $\sqrt{\cdot}$ (which corresponds to using the Hellinger's kernel in the SVM). In all cases, the descriptor $\mathbf{d}^*$ is a real vector that *has the same number of elements as a single video frame*. Therefore, $\mathbf{d}^*$ can be interpreted as a standard RGB image. Furthermore, since this image is obtained by rank pooling the video frames, it summarizes information from the whole video sequence.

A few examples of dynamic images are shown in fig. 1. Several observations can be made. First, interestingly the dynamic images tend to focus mainly on the acting objects, such as humans or horses in the "horse racing" action, or drums in the "drumming" action. On the contrary, background pixels and background motion patterns tend to be averaged away. Hence, the pixels in the dynamic image tend to focus on the identity and motion of the salient actors in videos, suggesting that they may contain the information needed to perform action recognition.

Second, we observe that dynamic images behave differently for actions of different speeds. For slow actions, like "blowing hair dry" in the first row of fig. 1, the motion seems to be dragged over many frames. For faster actions, such as "golf swing" in the second row of fig. 1, the dynamic image reflects key steps of the action such as preparing to swing and stopping after swinging. For longer term actions such as "horse riding" in the third row of fig. 1, the dynamic image reflects different parts of the video; for instance, the rails that appear as a secondary motion contributor are superimposed on top of the horses and the jockeys who are the main actors. Such observations were also made in [18].

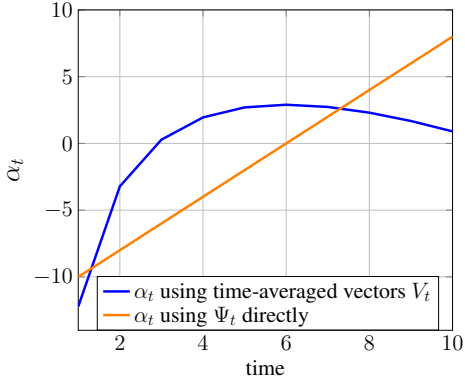

Fig. 3: The graph compares the approximate rank pooling weighting functions $\alpha_t$ (for $T = 10$ samples) of eq. (3) using time-averaged feature frames $V_t$ to the variant eq. (2) that ranks directly the feature frames $\psi_t$ as is.

Last, it is interesting to note that dynamic images are reminiscent of some other imaging effects that convey motion and time, such as *motion blur* or *panning*, an analogy is illustrated in fig. 2. While motion blur captures the time and motion by integrating over subsequent pixel intensities defined by the camera shutter speed, dynamic images capture time by integrating and reordering the pixel intensities over time within a video.

### 3.2 Fast dynamic image computation

Computing a dynamic image entails solving the optimization problem of eq. (1). While this is not particularly slow with modern solvers, in this section we propose an approximation to rank pooling which is much faster and works as well in practice. Later, this technique, which we call *approximate rank pooling* (ARP), will be critical in incorporating rank pooling in intermediate layers of a deep CNN and to allow back-prop training through it.

The derivation of ARP is based on the idea of considering the first step in a gradient-based optimization of eq. (1). Starting with $\mathbf{d} = \vec{0}$, the first approximated solution obtained by gradient descent is $\mathbf{d}^* = \vec{0} - \eta \nabla E(\mathbf{d})|_{\mathbf{d}=\vec{0}} \propto -\nabla E(\mathbf{d})|_{\mathbf{d}=\vec{0}}$ for any $\eta > 0$, where

$$\nabla E(\vec{0}) \propto \sum_{q>t} \nabla \max\{0, 1 - S(q|\mathbf{d}) + S(t|\mathbf{d})\}|_{\mathbf{d}=\vec{0}}$$
$$= \sum_{q>t} \nabla \langle \mathbf{d}, V_t - V_q \rangle = \sum_{q>t} V_t - V_q.$$

We can further expand $\mathbf{d}^*$ as follows

$$\mathbf{d}^* \propto \sum_{q>t} V_q - V_t = \sum_{t=1}^T \beta_t V_t$$

where $\beta_t$ are scalar coefficients. By expanding the sum

$$\sum_{q>t} V_q - V_t = (V_2 - V_1)$$
$$+ (V_3 - V_1) + (V_3 - V_2)$$
$$\vdots$$
$$+ (V_T - V_1) + (V_T - V_2) + \ldots + (V_T - V_{T-1}).$$

one can simply see that each $V_t$ with positive or negative sign occurs $(t-1)$ and $(T-t)$ times respectively. Now we can write $\beta_t$ in terms of time and video length:

$$\beta_t = (t-1) - (T-t) = 2t - T - 1. \quad (2)$$

The time average vectors $V_t$ can be written in terms of feature vectors $\psi_t$ and $\mathbf{d}^*$ can be written as a linear combination of $\psi_t$

$$\mathbf{d}^* \propto \beta_t V_t = \alpha_t \psi(I_t)$$

where the coefficients $\alpha_t$ are given by

$$\alpha_t = 2(T - t + 1) - (T + 1)(H_T - H_{t-1}), \quad (3)$$

where $H_t = \sum_{i=1}^t 1/t$ is the $t$-th Harmonic number and $H_0 = 0$. The $\alpha_t$ coefficients can be derived from the observation that each $\psi_t$ occurs $\sum_{i=t}^T \beta_i H_i$ times in the sum. Hence the rank pooling operator reduces to

$$\hat{\rho}(I_1, \ldots, I_T; \psi) = \sum_{t=1}^T \alpha_t \psi(I_t). \quad (4)$$

which is a weighted combination of the data points ($\psi(I_t)$). In particular, the dynamic image computation reduces to accumulating the video frames after pre-multiplying them by $\alpha_t$. The function $\alpha_t$ is illustrated in fig. 3.

An alternative construction of the rank pooling does not compute the intermediate average features $V_t = (1/t) \sum_{q=1}^T \psi(I_q)$, but uses directly individual video features $\psi(I_t)$ in the definition of the ranking scores (1). In this case, the derivation above results in a weighting function of the type

$$\alpha_t = 2t - T - 1 \quad (5)$$

which is linear in $t$. The two scoring functions eq. (3) and eq. (2) are compared in fig. 3 and in the experiments.

## 4 DYNAMIC MAPS NETWORKS

In the previous section we have introduced the concept of dynamic image as a method to pool the information contained in a number of video frames into a single RGB image. Here, we notice that every layer of a CNN produces as output a *feature map* which, having a spatial structure similar to an image, can be used in place of video frames in this construction. We call the result of applying rank pooling to such features a *dynamic feature map*, or *dynamic map* in short. In the rest of the section we explain this construction can be incorporated in a CNN as a rank-pooling layer (section 4.1) and how ARP can be used to accelerate it significantly as well as to perform back-propagation for end-to-end learning (section 4.2).

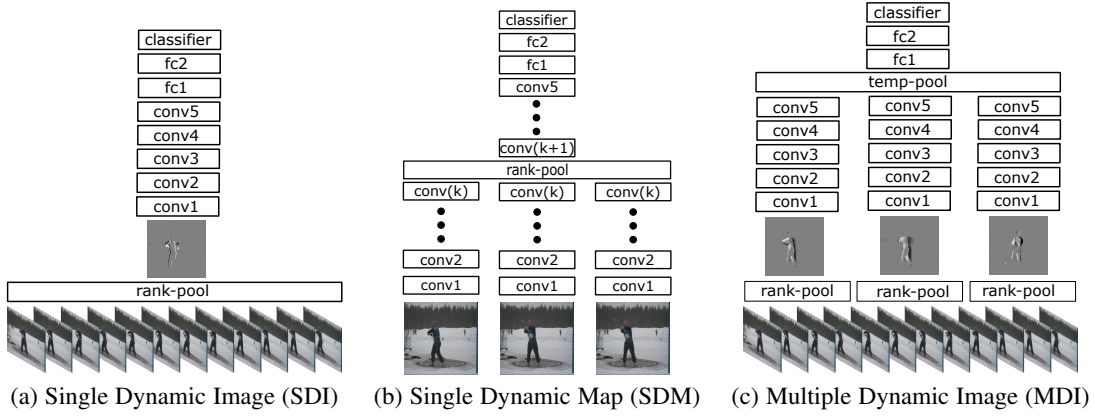

Fig. 4: Illustration of various dynamic image/map network architectures.

## 4.1 Dynamic maps

We illustrate three architecture designs for dynamic map networks fig. 4. In the case seen so far (fig. 4.(a)), rank pooling is applied at the level of the input RGB video frames, which can be though of as "layer zero" in the architecture. We call this architecture a *dynamic image network*. By contrast, a *dynamic map network* (fig. 4.(b)) moves rank pooling higher in the hierarchy, by applying one or more layers of feature computations to the individual feature frames and applying rank pooling to the resulting feature maps.

In particular, let $\mathbf{a}_1^{(l-1)}, \ldots, \mathbf{a}_T^{(l-1)}$ denote the feature maps computed at the $l-1$ layers of the architecture, one for each of the $T$ video frames. Then, we use the rank pooling equation (1) to aggregate these maps into a single dynamic map,

$$\mathbf{a}^{(l)} = \rho(\mathbf{a}_1^{(l-1)}, \ldots, \mathbf{a}_T^{(l-1)}). \tag{6}$$

Note that, compared to eq. (1), we dropped the term $\psi$; since networks are already learning feature maps, we set this term to the identity function. The dynamic image network is obtained by setting $l = 1$ in this construction.

**Rank pooling layer (RankPool) & backpropagation.** In order to train a CNN with rank pooling as an intermediate layer, it is necessary to compute the derivatives of eq. (6) for the backpropagation step. We can rewrite eq. (6) as a linear combination of the input data $V_1, \ldots, V_T$, namely

$$\mathbf{a}^{(l)} = \sum_{t=1}^{T} \beta_t(V_1, \ldots, V_T) V_t \tag{7}$$

In turn, $V_t$ is the temporal average of the input features and is therefore a linear function $V_t(\mathbf{a}_1^{(l-1)}, \ldots, \mathbf{a}_t^{(l-1)})$. Substituting, we can rewrite $\mathbf{a}^{(l)}$ as

$$\mathbf{a}^{(l)} = \sum_{t=1}^{T} \alpha_t(\mathbf{a}_1^{(l-1)}, \ldots, \mathbf{a}_T^{(l-1)}) \mathbf{a}_t^{(l-1)}. \tag{8}$$

Unfortunately, we observe that due to the non-linear nature of the optimization problem of equation (1), the coefficients $\beta_t, \alpha_t$ depend on the data $\mathbf{a}_t^{(l-1)}$ themselves. Computing the gradient of $\mathbf{a}^{(l)}$ with respect to the per frame data points $\mathbf{a}_t^{(l-1)}$ is a challenging derivation. Hence, using dynamic maps and rank pooling directly as a layer in a CNN is not straightforward. This problem is solved in the next section.

We note that the rank pooling layer (RankPool) constitutes a new type of portable convolutional network layer, just like a max-pooling or a ReLU layer. It can be used whenever dynamic information must be pooled across time.

## 4.2 Approximate dynamic maps

Due to the intrinsic noisy nature of image and video data, computing dynamic images/maps to a high degree of accuracy may not be necessary in practice. In fact, accurate optimization of eq. (6) has two disadvantages: optimization is slow and computing the derivative for backpropagation is difficult. This is especially important in the context of CNNs, where efficient computation and end-to-end learning is extremely important for training on large datasets.

To this end we replace once again rank pooling with ARP. ARP significantly accelerates the computations, up to a factor of 45 as we show later in the experiments. Furthermore, and more importantly, the ARP is a linear combination of frames, where the per frame coefficients are given by eq. (3). These coefficients are independent of the frame features $V_t$ and $\psi(I_t)$. Hence, the derivative of ARP is simpler and fast to compute:

$$\frac{\partial \operatorname{vec} \mathbf{a}^{(l)}}{\partial (\operatorname{vec} \mathbf{a}_t^{(l-1)})^\top} = \alpha_t I \tag{9}$$

where $I$ is the identity matrix and $\operatorname{vec}$ denotes the tensor stacking operator [31]. Formally, the same expression can be obtained by computing the derivative of eq. (8) pretending that the coefficients $\alpha_t$ do not depend on the video frames.

We conclude that using ARP in the context of CNNs speeds up evaluation and dramatically simplifies optimization through backpropagation.

## 4.3 Single and multiple dynamic map networks

Dynamic images and maps can be computed over an arbitrary length video sequences. Here we propose a few deep network variants that can use a single or multiple dynamic images or maps integrated over different video sequence durations.

**Single Dynamic Image/Map (SDI/SDM).** In the first scenario, a dynamic image/map summarizes an entire video sequence. By training a CNN on top of such dynamic images, the method implicitly captures the temporal patterns contained in the video. However, since the CNN is still applied to images, we can start from a CNN *pre-trained for still image recognition*, such as CaffeNet pre-trained on the ImageNet ILSVRC data, and fine-tune it on a dataset of dynamic images. Fine-tuning allows the CNN to learn features that capture the video dynamics without the need to train the architecture from scratch. This is an important benefit of our method because training large CNNs require millions of data samples which may be difficult to obtain for videos.

**Multiple Dynamic Images/Maps (MDI/MDM).** While fine-tuning requires less annotated data than needed for training a CNN from scratch, the domain gap between natural and dynamic images is sufficiently large that fine-tuning may still requires a relatively large annotated dataset. Unfortunately, as noted above, in most cases there are only a few videos available for training.

In order to address this potential limitation, in the second scenario we propose to generate multiple dynamic images/maps from each video by breaking it into segments. In particular, for each video we extract multiple partially-overlapping segments of duration $\tau$ and with stride $s$. In this manner, we create multiple video segments per video, essentially multiplying the dataset size by a factor of approximately $T/s$, where $T$ is the average number of frames per video. This can also be seen as a data augmentation step, where instead of mirroring, cropping, or shearing images we simply take a subset of the video frames. From each of the new video segments, we can then compute a dynamic image/map to train the CNN, using as ground truth class information of each subsequence the class of the original video.

The use of multiple, shorter subsequences also reduces the amount of temporal information that is squeezed in a single dynamic image or map, which can be beneficial in modeling highly-complex videos with many temporal changes.

The resulting network architecture (fig. 4.(c)) takes a sequence of frames from a video as input and splits them into fixed length subsequences, generating a dynamic image/map for each one subsequence. The last convolutional layer is followed by a "temporal pooling" layer which merges the dynamic images/maps into one. We evaluate different choices for this temporal pooling layer in the experiments section. Note that, while (fig. 4.(c)) show the case of a multiple dynamic image (MDM) network, the figure is easily adapted to a multiple dynamic map (MDM) by moving the approximate rank pooling layer at higher layers.

### 4.4 Four-stream architecture

The concept of dynamic image can be applied to different video input modalities such as depth and optical flow data. As the combination of RGB and optical flow has been shown to be very useful in action recognition in the two-stream architecture of Simonyan and Zisserman [57], we experiment here with a similar idea and propose a new four-stream architecture for action recognition (fig. 5). As for the two-stream model, this architecture combines RGB and optical flow data streams, either directly or by computing dynamic images/maps. This means that the network processes static appearance and visual context information from the RGB stream, low-level motion information from the optical flow stream, mid-level motion information from dynamic images computed from RGB data (dynamic image stream), and higher-level motion information from dynamic images computed from optical flow data (dynamic optical flow stream).

## 5 EXPERIMENTS

In this part, we first give details for the video classification benchmarks (section 5.1) and experimental setup (section 5.2) used in the paper. Then, we thoroughly evaluate and compare the SDI and MDI architectures, ARP, dynamic maps and parametric pooling. Finally we show that our results are on par and complementary to the current state-of-the-art.

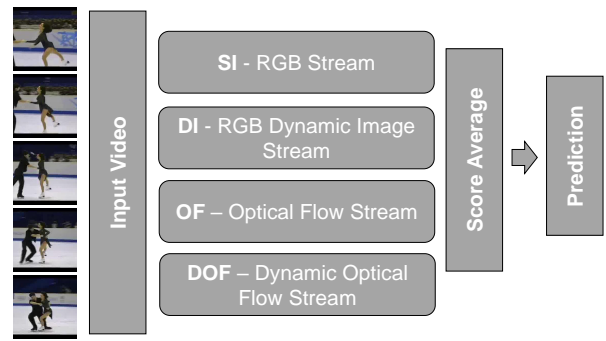

Fig. 5: The illustration of four stream dynamic image architecture that combines RGB data, Optical Flow with Dynamic Images and Dynamic Optical Flow.

### 5.1 Datasets

We benchmark on two popular datasets for action recognition: *UCF101* [60] and HMDB51 [33].

**UCF101.** The UCF101 dataset [60] comprises of 101 human action categories, like "Apply Eye Makeup" and "Rock Climbing" and contains over $13,320$ videos. The videos are realistic and relatively clean. They contain little background clutter, a single action label, and are trimmed around that action (thus almost all frames relate to the labelled action). Performance is evaluated in term of average recognition accuracy over three data splits provided by the authors.

**HMDB51.** The HMDB51 dataset [33] consists of 51 human action categories, such as "backhand flip" and "swing baseball bat" and spans over $6,766$ videos. The videos are realistic (downloaded from Youtube) each containing a single human action. This dataset is split in three parts and accuracy is averaged over all three parts, similar to *UCF101*.

### 5.2 Implementation details

This section describes the details of the models used in the experiment. Full source code and models using the MatConvNet toolbox [66] are available online.[2].

**Network:** We use two deep neural network architectures. The first is the BVLC reference model (CaffeNet) [28] which is reasonably efficient to train and evaluate. We use this model to analyze various design decisions such as different pooling methods and their point of application in the network. After identifying the most promising settings, we use them with ResNeXt-50 and ResNeXt-101 [75] (using the $32 \times 4d$ variant). All models are are pretrained on ImageNet ILSVRC 2012 [53] and fine-tuned by using stochastic gradient descent. During training, we randomly flip images, jitter image size and aspect ratio and rescale it to $224 \times 224$. In test time, we use a single center crop for each dynamic image.

**RGB and optical flow:** We take each video and convert it into frames at its original frame rate. In addition to the extracted RGB frames, we also precompute optical flow using the method of [79] and store the flow fields as JPEG images after clipping displacement to 20 pixel and rescaling the flow values in the range 0 to 255.

---
[2]. https://github.com/hbilen/dynamic-image-nets

| Method | Accuracy (%) |
|---|---|
| Mean Image | 52.6 |
| Max Image | 48.0 |
| Motion History Image | 46.6 |
| Dynamic Image | **57.2** |

TABLE 1: Comparing the performance of various single image video representation methods on split-1 the UCF101 dataset in terms of mean class accuracy (%).

### 5.3 Rank pooling

**Max and average pooling.** First, we evaluate the "single dynamic image" (SDI) setting (section 4.3), namely extracting a single dynamic image for each video sequence. We compare SDI with the two most popular pooling methods, namely mean and max pooling, obtaining alternative video summary images. All pooling methods are applied offline and the resulting summary images are cached. Dynamic images, in particular, are computed using the SVR software of [58]. Then, the CaffeNet model (see fig. 4.(a)) is fine-tuned for each summary image variant using the first train/test split of the UCF101 dataset.

In order to generate dynamic images, we follow the pipeline suggested in Fernando *et al.* [17]: i) square root the RGB pixel values ($\psi(\cdot)$), ii) use a time varying mean representation of $\sqrt{\cdot}$, iii) learn ranking hyperplanes for each channel, iv) scale the computed dynamic images into $[0, 255]$ range again. The dynamic images are precomputed and fed into a CNN as input in the experiments.

Results are shown in table 1. We observe that DIs achieve the highest accuracy and conclude that rank pooling is preferable to mean and max pooling for video summarization.

**Motion History Images (MHI).** MHI [3] is a direct competitor to our dynamic image as it also generates a single image summarizing a video. An MHI is a scalar-valued image built such that pixels that changed more recently in the video are brighter, so that pixel intensity is a function of the changes observed at that point. A qualitative comparison between dynamic images (generated using ARP) and MHIs in fig. 6. The figure shows, from top to bottom, a representative frame for a given video and the corresponding MHI and DI. We first note that DIs provide more detailed representation of the videos, as the range of intensity values are not limited with the number frames as in MHIs. Second, DIs are more robust to moving viewpoint and background. Finally, MHIs can only represent the motion gradient in object boundaries in contrast to DIs.

MHIs were originally designed for action recognition. A set of moment-based descriptors are extracted from a set of MHIs, then a distance metric over each action category is learnt and classification is performed using the computed metrics. Such a pipeline is not competitive for the modern datasets, and, thus, we adapt it to fit modern pipelines. Similar to our method, we compute a single MHI for each video, which we use as input to the CaffeNet model and train on the UCF101 dataset. For this representation, we obtain $46.6\%$ accuracy in the first split of the UCF101 dataset which is significantly lower ($-10\%$) than what we obtain with SDI. This suggests that the qualitative advantages translate in better quantitative classification performance as well.

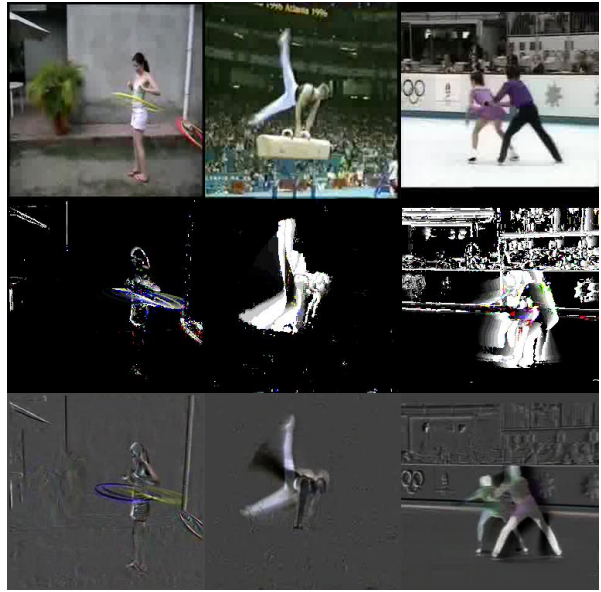

Fig. 6: Comparing Dynamic Images (DI) to Motion History Images (MHI) [3]. The top row shows representative frames from different videos, middle and bottom rows depict MHI and DI of corresponding videos respectively. While both methods can represent the evolution of pixels along time, our method produces more interpretable images which are more robust to long-range and background motion.

| Method | fps | Ranking Acc. | Classification Acc. |
|---|---|---|---|
| RP | 131 | $95.5 \pm 0.6$ | 57.9 |
| ARP | 5920 | $96.5 \pm 0.5$ | 55.2 |

TABLE 2: Approximate rank pooling vs rank pooling in terms of speed, ranking accuracy (%) and classification accuracy (%)

### 5.4 Approximating rank pooling

Next, we compare the rank pooling (RP) to approximate rank pooling (ARP) in terms of speed (frames per second) and pairwise ranking accuracy, which is a common measure for evaluating learning-to-rank methods. Hence the goal is to assess the ability of ARP to sort video frames correctly compared to its "exact" counterpart RP. More importantly, we also compare RP and ARP in terms of overall recognition performance to see if the approximation impacts the ability of the method to represent video effectively.

To do so, we apply RP and ARP to the videos from the first test split of UCF101 dataset which contains 3783 video-clips in varying lengths. We report results with the mean and the standard deviations in table 2. Interestingly, ARP achieves slightly better pairwise ranking accuracy than RP (96.5% vs 95.5%). This can be attributed to the fact that ARP optimizes the actual target pairwise ranking loss (see eq. (1)), while the Support Vector Machine Regression solves a regression problem from RGB values to frame index. Although both pooling obtain high ranking performances overall, we observe that both method have relatively lower performance to correctly rank frames from categories with periodic motion and static background such as "PushUp" (82.5%), "JugglingBalls" (92.7%) and "PlayingGuitar" (92.7%) have lower ranking accuracies.

In terms of run-time, RP takes a second to learn a dynamic image from a 150 frame-length video in average. Since this is approximately five times slower than a forward pass of the CNN model (CaffeNet), RP slows down the system considerably. ARP is $\sim 45\times$ faster than RP, hence adding negligible cost to the CNN computation, while obtaining comparable ranking performance. The quality of the approximation is corroborated by fig. 7 which shows that the ranking score distributions for RP and ARP are also similar while ARP achieves slightly better results.

| window | stride | Accuracy |
|---|---|---|
| 5 | 3 | 63.9 |
| 10 | 3 | 66.9 |
| 10 | 6 | **67.4** |
| 20 | 6 | 60.7 |
| 20 | 12 | 58.9 |
| video length | - | 55.2 |

TABLE 3: MDI: Effect of window size ($\tau$) and stride ($s$) in terms of multi-class classification accuracy in the first split of the UCF101.

| depth | dynamic image conv1 | dynamic map conv2 | conv3 | conv4 |
|---|---|---|---|---|
| UCF101 | 70.9 | 68.5 | **73.3** | 67.6 |
| HMDB51 | 37.2 | **38.0** | **38.0** | 37.4 |

TABLE 4: Classification accuracy (%) for dynamic image and map networks at different depths on the UCF101 and HMDB51 datasets. "conv1" corresponds to placing the ARP as the first layer of network.

We also compare RP and ARP in terms of classification accuracy on the first split of the UCF101 dataset. Here the single dynamic image representation with ARP obtains $55.2\%$ accuracy which is 2.7 points lower than RP. ARP is however much faster than RP and still significantly outperforms mean (52.6 points) and max pooling (48 points).

Due to the excellent accuracy and speed, in the rest of the paper we will use by default approximate dynamic images, computed with using ARP instead of RP, unless otherwise noted.

### 5.5 Single vs multiple dynamic image networks

So far, we used a single dynamic image to represent each video using the CNN architecture of fig. 4.(a). In this section we show that splitting a video into multiple sub-sequences and encoding each of them as a dynamic image achieves better classification accuracy than using a single dynamic image. To do so, we use the multiple dynamic image (MDI) network of fig. 4.(c). After applying the ARP to each sub-sequences and extracting sequence-specific features, this CNN uses an additional temporal max-pooling layer to merge the resulting convolutional features into one (denoted `temp-pool` in fig. 4.(c)).

First, we evaluate the effect of window size ($\tau$) and stride ($s$) which determine the sub-sequence length and frame sampling frequency, respectively (see section 4.3 for the definition of such parameters). Note that using video-length window size is equivalent to computing a single dynamic image per video and a single frame window corresponds to using RGB frames as input. As shown in table 3, using a medium-length windows of 10 samples with $40\%$ overlap yields the best classification performance. However, while MDI is more accurate than SDI, it is also slower due to the fact that MDI extract features from a number of dynamic images proportional to the video length. In practice, MDI is 5 times slower than SDI for a medium-length video, so SDI can be preferable when speed is paramount.

Figure 8 shows several dynamic images computed by varying window sizes (from top to bottom: 10 samples, 50 samples and whole video length). As more frames are used to generate dynamic images, more pixels are activated to represent longer motions. For instance, in fig. 8.(a) and (b) using longer windows results in images that capture more revolutions of wheels and hula-hoops. We also notice that the dynamic image representation fails to capture very complex motion in fig. 8.(c) as the number of frames increases too much.

Finally, we evaluate different choices for the temporal pooling layer `temp-pool`. Using mean pooling, max pooling and APR for this layer results in 66.2, 68.3 and 65.2% mean video classification accuracies respectively. The fact that max pooling gives the best result can be explained with the fact that max pooling is invariant to the temporal position of the target action instance and does not require any alignment of start and end of action instances across different videos. This is in contrast with encoding shorter video sequences, where we demonstrated that ARP is better than both sum and max pooling.

### 5.6 Dynamic maps

So far we used ARP as the first layer of the CNN to generate dynamic images. Here we move ARP deeper down the network to generate dynamic maps instead. Table 4 reports the mean class accuracy on the UCF101 and HMDB51 datasets, where "convX" corresponds to positioning the approximate rank before the $X$-th convolutional layer. For instance "conv1" means that ARP is at applied directly to the input images. Each network is trained in an end-to-end manner with multiple dynamic maps (see the network architecture in fig. 4.(b)). Please note that such an end-to-end training is only possible because ARP enables backpropagation, which would be difficult to do with RP (section 4.2).

We see that locating ARP before "conv3" performs slightly better than "conv1" and "conv2" and the classification performance starts degrading after this level. The degradation can be explained with the fact that the convolutional features in the earlier layers, which capture low-level structures such as edges, blobs and patterns, are more useful to express the motion and dynamics of a video.

### 5.7 Parametric pooling

As shown in section 4.2, ARP is a fixed linear combination of input images where the mixing coefficients $\alpha_t$ are given by the formula eq. (4) derived from the ranking objective. A natural question is whether better coefficients $\alpha_t$ could be obtained by optimise the target task of video classification end-to-end instead of using ranking as a proxy task. We call this setting *parametric pooling*. Similarly to ARP, parametric pooling takes a number of frames or feature tensors from a video as input and pools them into a single frame/feature tensor. In contrast to ARP, for which eq. (4) applies to videos of any length, parametric pooling requires videos of a fixed length.

In practice, parametric pooling can be implemented as a sub-network which is composed of a number of convolutional layers

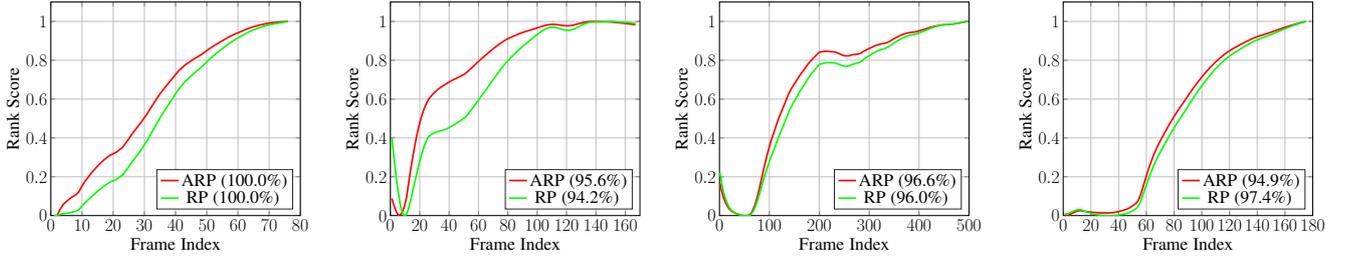

Fig. 7: Comparison between four score profiles and pairwise ranking accuracies (%) of ranking functions for approximate rank pooling (ARP) and rank pooling (RP). Generally the approximate rank pooling follows the trend of rank pooling.

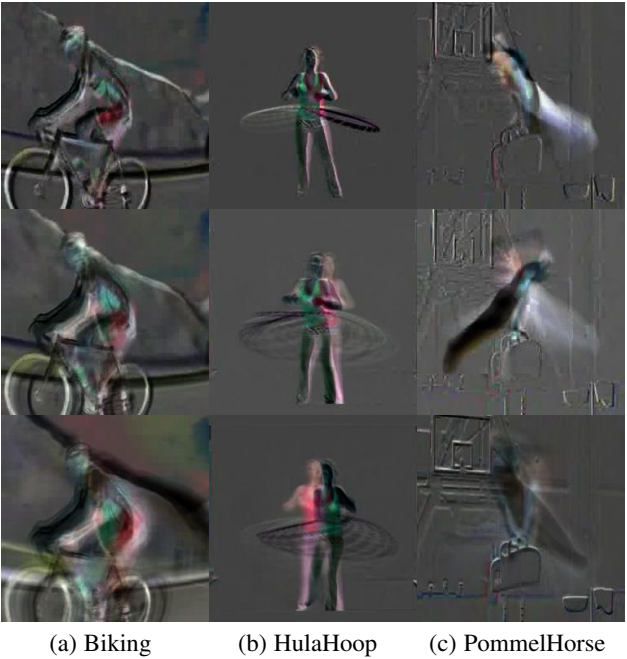

(a) Biking  (b) HulaHoop  (c) PommelHorse

Fig. 8: Visual analysis of different window sizes ($\tau$) on dynamic images. The top, middle and bottom rows depict dynamic images for $\tau = 10$, $\tau = 50$ and $\tau =$(whole video length) respectively. Best seen in color.

| pooling | dataset | conv1 | conv2 | conv3 | conv4 |
|---|---|---|---|---|---|
| ARP | UCF101 | 70.9 | 68.5 | 73.3 | 67.6 |
|  | HMDB51 | 37.2 | 38.0 | 38.0 | 37.4 |
| PP1 | UCF101 | 67.6 | 73 | 71.9 | 70.9 |
|  | HMDB51 | 34.7 | 36.6 | **38.4** | 37.4 |
| PP2 | UCF101 | 68.5 | **73.9** | 70.0 | 69.1 |
|  | HMDB51 | 36.0 | 36.9 | 37.3 | 37.4 |

TABLE 5: Classification accuracy (%) for approximate rank pooling and parametric pooling at different depths on the UCF101 and HMDB51 datasets. "conv1" corresponds to placing the ARP or PPX as the first layer of network. PP1 and PP2 correspond to 1 and 2 layer sub-networks respectively.

followed by non-linear operators. In case of a single convolution and ReLU layers, it can be formulated as

$$\phi(I_1,\ldots,I_T;\psi) = \sigma\left(\sum_{t=1}^{T} \alpha_t \psi(I_t) + b_t\right)$$

where $\sigma$ is the ReLU function and $b_t$ is a bias parameter. Scalar multiplication and sum over time by $(\alpha_t, b_t)$ can be interpreted as a fully-connected layer with one scalar output applied to temporal slices of the data tensors. This can be extended to use several temporal layers, as we do next.

In table 5 we evaluate the performance of the proposed parametric pooling on the UCF101 and HMDB51 datasets for two settings PP1 and PP2, using one or two layers in the parametric pooler. Similar to the dynamic map experiments, we apply parametric pooling after the ReLU layer following the specified convolutional layer and parametric pooling is followed by an additional ReLU. The parameters of the networks are trained in end-to-end fashion with the pooling coefficients. The single layer parametric pooling (PP1) is implemented as a $10 \times 1$ temporal convolutional layer over 10 frames that belong to the same video. This also means that the number of frames is fixed to 10 for each video subsequence (see table 3). PP2 extend PP1 by considering a chain of $10 \times 10$ and $10 \times 1$ temporal fully-connected layers (with ReLU in between). Table 5 shows that PP1 and PP2 performance is similar to ARP with the exception that parametric pooling performs worse on the raw video frames ("conv1").

### 5.8 Dynamic images with deeper networks

In the previous experiments we have used the fast CaffeNet architecture to explore certain design decisions. Next, we evaluate action recognition with dynamic images using more recent networks such as ResNeXt-50 [75]. Results are reported in table 6.

We make several observations. First, switching from CaffeNet to ResNeXt-50 boosts performance significantly, up to 15% for UCF101 and 20% for HMDB51. This is in line with the top-1 error rate in the validation split of the ImageNet dataset *i.e.* 42.6 and 22.6% for CaffeNet and ResNeXt-50 respectively. Second, SI and DI streams are highly complementary both for CaffeNet and ResNeXt-50. Third, while for UCF101 the performances of SI and DI are on par, for HMDB51 the DI stream alone scores considerably higher (4%). The reason is that in UCF101 many videos can be recognized from only the static context and background, while in HMDB51 backgrounds are more complex and dynamic.

We further break down the comparison of static and dynamic image networks on a per-class basis. In order to do so, we compute the top 3 classes based on the relative performances for SI and DI. DI performs better for "Nunchucks", "Jumping-Jack", "WallPushups", where longer term motion is dominant and discriminating between motion patterns is important. SI works better for classes such as "HammerThrow", "Shotput" and "BreastStroke", where context is already quite revealing (*e.g.* swimming pool for "BreastStroke") and dynamics are not enough themselves to distinguish an action type from another (*e.g.* DI confuses "BreastStroke" with "FrontCrawling" and "Rowing"). We conclude that dynamic images are useful recognition of actions with characteristic motion patterns.

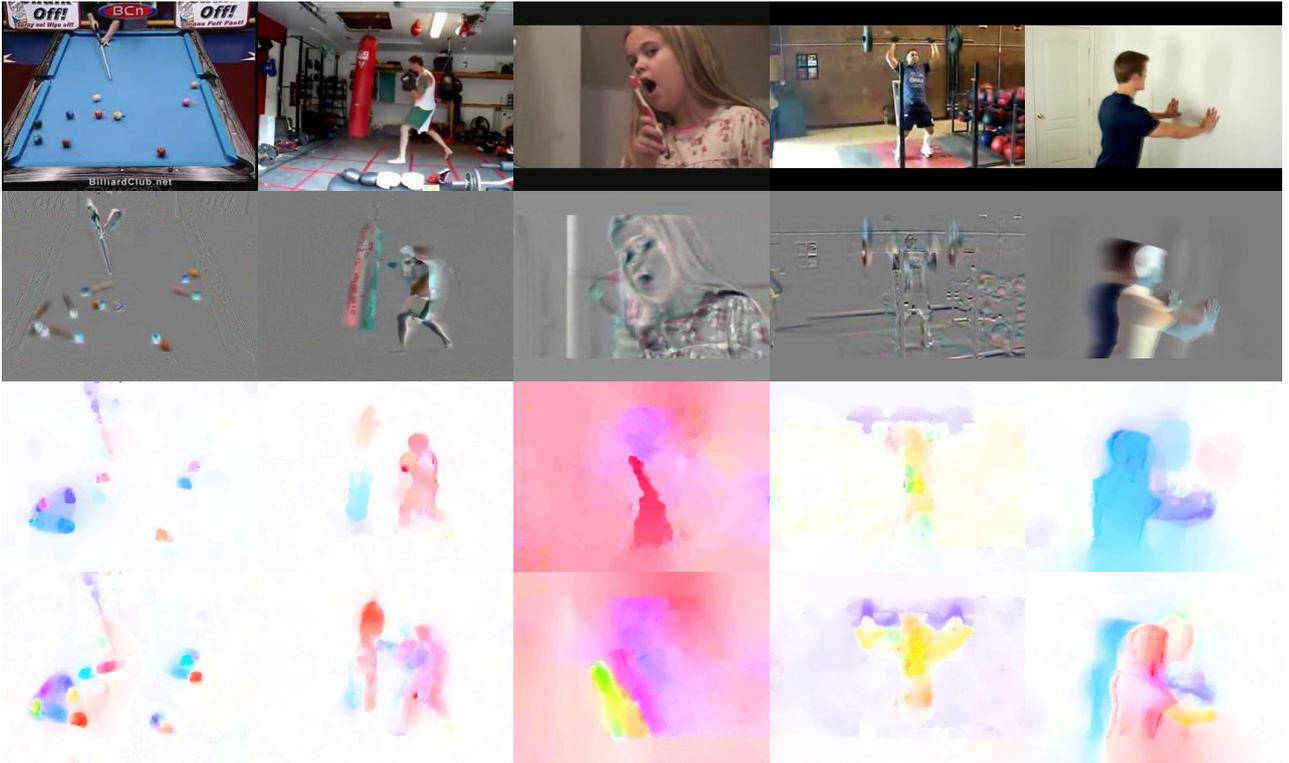

Fig. 9: Visualizing static images (SI), dynamic images (DI), optical flow (OF) and dynamic optical flow (DOF) in each row respectively. Best seen in color.

|  |  | SI | DI | SI+DI |
|---|---|---|---|---|
| UCF101 | CaffeNet [28] | 68.5 | 70.9 | 76.5 |
| UCF101 | ResNeXt-50 [75] | 87.6 | 86.6 | 90.6 |
| HMDB51 | CaffeNet [28] | 36.0 | 37.2 | 39.7 |
| HMDB51 | ResNeXt-50 [75] | 53.5 | 57.3 | 61.3 |

TABLE 6: Classification accuracy (%) with dynamic images when using CaffeNet [28] and deeper convolutional network architectures, specifically ResNeXt-50 [75]. As we can observe, dynamic images can reap all the benefits of deeper architectures of modern convolutional neural networks.

### 5.9 Dynamic optical flow

So far, we use RGB frames as input to our dynamic image/map networks. Inspired by the success of two stream networks [57] which combine both RGB and optical flow images, we extend our models to include optical flow images as well. This, in a similar fashion to the static image (SI) and dynamic image (DI) networks, we introduce optical flow (OF) and dynamic optical flow (DOF) streams in our model. DOF is obtained by applying ARP to 10 optical flow frames. Differently from DI, a DOF contains two channels (corresponding to horizontal and vertical flow) rather than 3 RGB channels. Figure 9 shows samples from different videos and action categories for SI, DI, OF and DOF streams. We observe two main differences between the raw optical flow and dynamic optical flow samples (third and last rows respectively). We first see that DOF can capture longer-term motion than OF.

|  |  | OF | DOF | OF+DOF |
|---|---|---|---|---|
| UCF101 | ResNeXt-50 [75] | 84.9 | 86.6 | 89.1 |
| HMDB51 | ResNeXt-50 [75] | 55.8 | 58.9 | 62.6 |

TABLE 7: Optical flow and dynamic optical flow streams: A two-stream ResNeXt-50 architecture for action classification in terms of mean multi-class accuracy (%).

This is expected as optical flow by definition captures the motion information between only subsequent frames. For instance, DOF can represent longer temporal history for the billiard balls and a longer span for a "punching" action (see in fig. 9). Second DOF can also represent higher order statistics such as "velocity" of optical flow. In the examples, one can note the forward acceleration of the boxer and of the toothbrush and the upward acceleration of the weightlifter.

Table 7 compares the action classification performance using ResNeXt-50 for the OF and DOF streams as well as their their combination. First, we note that combining optical flow with dynamic optical flow improves the performance of the individual streams, confirming that the two features are complementary. Second, ARP works as well for optical flow image as for RGB images. In fact, dynamic optical flow alone achieves a very high accuracy on UCF101 (86.6%). Last we show that OF and DOF streams are complimentary and using two-stream OF and DOF leads to 89.1% and 62.6% in the UCF101 and HMDB51 and obtains significant improvement over the individual streams.

Next we break down our analysis on a per-class basis, focusing on the top 3 classes with the highest relative performances for OF and DOF. While action categories characterised by longer term motion and higher order statistics such as "Nunchucks", "HandstandWalking" and "JumpingJack" are the best for DOF, actions characterised by shorter motions such as "BreastStroke", "HighJump", "BlowDryHair" are best for OF.

|  | HMDB51 | UCF101 |
| --- | --- | --- |
| SI | 53.5 | 87.6 |
| DI | 57.3 | 86.6 |
| OF | 55.8 | 84.9 |
| DOF | 58.9 | 86.6 |
| SI+OF | 67.5 | 93.9 |
| SI+DI | 61.3 | 90.6 |
| OF+DOF | 62.6 | 89.1 |
| SI+DI+OF+DOF | 71.5 | 95.0 |

TABLE 8: Combinations of various RGB and optical flow streams with ResNeXt-50 in terms of mean multi-class accuracy (%).

### 5.10 Four stream networks

Next, we evaluate the combinations of SI, DI, OF and DOF streams. While these networks are trained individually, at test time their outputs (or classification scores) are simply averaged to obtain an overall score. We show the results of different stream combinations in table 8. First we see that combining SI and DI even without using any optical flow achieves significant improvement (7.8 and 4 points in HMDB51 and 3 and 4 points in UCF101) over individual SI and DI streams respectively. Similarly DOF is complementary to the OF stream, their combination leads to 62.6% and 89.1% in HMDB51 and UCF101 resp. Finally, combining all the four streams obtains remarkable classification accuracy, improving over all the two stream networks (4 and 1.1 points over SI+OF, 10 and 4.4 over SI+DI, 8.9 and 5.9 over OF+DOF). To demonstrate the significance of the improvements, we run independent two-sample t-tests for all the two stream combinations for RGB frames and optical flow (SI+DI, OF+DOF) and the four stream one (SI+DI+OF+DOF). The statistical test results validate that the improvements are statistically significant at 0.05 level.

Next, we break down the analysis on a per-class basis looking at the worst performing classes in the UCF101 dataset for the four-stream ResNeXt-50 model. The most challenging five categories for this model are "PizzaTossing", "Lunges", "HammerThrow", "ShavingBeard" and "BrushingTeeth" with the respective accuracies of 74, 74.6, 77.2, 78.7 and 80.2%. Hence the four-stream architecture may fail to distinguish action categories separated by subtle differences. For instance, "Lunges", "CleanAndJerk", "BodyWeightSquats" may all involve subactions like lifting or lowering a barbell and kneeing, and are mostly distinguished by the order between such subactions. A possible solution could be to discover such subactions during learning and model their order by using rank pooling. Other similar actions, such as "ShavingBeard", "BrushingTeeth" and "ApplyLipstick" that involve similar motions may be confused in some cases. Incorporating specialized networks for facial and human body pose analysis may help in such cases.

### 5.11 State-of-the-art comparisons

Table 9 depicts a quantitative comparison of our four-stream network (SI+DI+OF+DOF) to the state-of-the-art on the UCF101 and HMDB51. In addition to ResNeXt-50 model, here we also train our model with the deeper ResNeXt-101 [75] and report its performance as well. In order to provide a fair comparison, we split the table into two parts, the ones incorporate their methods with the handcrafted improved dense trajectories (iDT) [68] to improve their final accuracy, and those that do not.

| Method | UCF101 | HMDB51 |
| --- | --- | --- |
| CNN-hid6 [80] | 79.3 | – |
| Comp-LSTM [62] | 84.3 | 44.0 |
| C3D+SVM [65] | 85.2 | – |
| 2S-CNN [78] | 88.0 | 59.4 |
| FSTCN [63] | 88.1 | 59.1 |
| 2S-CNN+Pool [78] | 88.2 | – |
| Objects+Motion($R^*$) [26] | 88.5 | 61.4 |
| 2S-CNN+LSTM [78] | 88.6 | – |
| TDD [70] | 90.3 | 63.2 |
| Temporal Segment Networks [71] | 94.2 | 69.4 |
| Two-Stream I3D [4] | 93.4 | 66.4 |
| Two-Stream I3D+ [4] (Kinetics300k) | 98.0 | 80.7 |
| *Four-Stream with ResNeXt-50 (Ours)* | 95.0 | 71.5 |
| *Four-Stream with ResNeXt-101 (Ours)* | 95.5 | 72.5 |
| FV+IDT [48] | 84.8 | 57.2 |
| SFV+STP+IDT [48] | 86.0 | 60.1 |
| FM+IDT [47] | 87.9 | 61.1 |
| MIFS+IDT [35] | 89.1 | 65.1 |
| CNN-hid6+IDT [80] | 89.6 | – |
| C3D Ensemble+IDT (Sports-1M) [65] | 90.1 | – |
| C3D+IDT+SVM [65] | 90.4 | – |
| TDD+IDT [70] | 91.5 | 65.9 |
| Sympathy [9] | 92.5 | 70.4 |
| Two-Stream Fusion+IDT [15] | 93.5 | 69.2 |
| ST-ResNet+IDT [14] | 94.6 | 70.3 |
| *Four-Stream+IDT with ResNeXt-50 (Ours)* | 95.4 | 74.2 |
| *Four-Stream+IDT with ResNeXt-101 (Ours)* | 96.0 | 74.9 |

TABLE 9: Comparison with the state-of-the-art in terms of mean multi-class accuracy (%). Our method outperforms the state state-of-the-art. Please note that better performing Two-Stream I3D+ [4] has been pre-trained on a large-scale video dataset, Kinetics300k.

First we look at the ones without iDT and see that the proposed four-stream network obtains the highest accuracy with 95.4% and 96.0% with ResNeXt-50 and ResNeXt-101 respectively. We outperform the state-of-the-art methods with a significant margin with the exception of I3D+ [4] (98% and 80.7%). Note that this method is pre-trained on additional 300,000 videos and relies on a two-stream variant. When trained on the UCF101 and HMDB51 alone, the I3D is outperformed by our four-stream architecture (93.4% and 66.4%). In any case, the I3D architecture can also incorporate dynamic images and enjoy a further boost. Remarkably, our method using *only* static and dynamic images and no optical flow still scores an impressive 90.6%, outperforming most competitors who rely on handcrafted optical flow input.

The four-stream architecture outperforms all previous methods even after incorporating the improved trajectory technique. This is encouraging as most of the best existing methods require improved trajectories to reach state-of-the-art accuracies. Furthermore, our four stream models do not improve significantly after the inclusion of improved trajectories (95.5% → 96.0% and 72.5% → 74.9%), showing that the vast majority of the benefit is intrinsic to the proposed architecture. This is interesting, as our four stream models are one of the first models together with I3D [4] which manages to surpass the 95% and 70% barriers on respective UCF101 and HMDB51 without relying on handcrafted features.

## 6 CONCLUSION

We have introduced the concepts of *dynamic images* and *dynamic optical flow*, powerful and yet simple video representations that summarizes videos into single images. Dynamic images and dynamic optical flow are able to encode the gist of the video first and second order dynamics, allowing for excellent action recognition performance. As they effectively comprise inputs to models, they can be used with any of the existing or future CNN architectures. In fact, applying dynamic images and dynamic optical flow with recent very deep convolutional neural networks enables end-to-end video action recognition with excellent results. We, furthermore, introduce a novel temporal pooling layer called *approximate rank pooling*, which accelerates dynamic image computation, while generalizing the idea to any intermediate feature map computed by a CNN. Approximate rank pooling allows for allowing back propagation of the gradients for learning. Furthermore, we proposed a novel four-stream architecture that combines complementary static and dynamic information from RGB and optical flow frames. Experiments on public action recognition benchmarks clearly demonstrate the benefits of th four-stream architecture, computing dynamic images onr RGB and optical flow images and achieving impressive performance despite their implementation simplicity.

Dynamic images have some notable limitations as well. Even though they are good at capturing smooth dynamics, they are less good at handling abrupt changes in very complex video sequences. Second, appearance and dynamics are highly correlated in the spatial and temporal domain, and it could be more efficient to build representations after decorrelating spatio-temporal volumes. Third, dynamic images operate at a single level of temporal pooling with a fixed window size. In future we plan to explore applying dynamic pooling at multiple levels of abstraction by allowing the network to adapt according to the complexity of temporal data. Furthermore, it would be interesting to evaluate extending the representation to modalities other than RGB and optical flow, such as depth and multi-spectral video data.

## ACKNOWLEDGMENTS

This work acknowledges the support of the EPSRC grant EP/L024683/1, the ERC Starting Grant IDIU and the Australian Research Council Centre of Excellence for Robotic Vision (project number CE140100016).

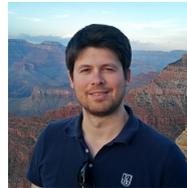

**Hakan Bilen** is a lecturer in the School of Informatics at the University of Edinburgh.

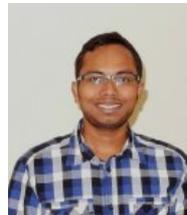

**Basura Fernando** is a research fellow with the The Australian Research Council Centre of Excellence for Robotic Vision based at the Australian National University.

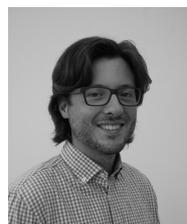

**Efstratios Gavves** is an assistant professor with the University of Amsterdam in the Netherlands and Scientific Manager of the QUVA Deep Vision Lab.

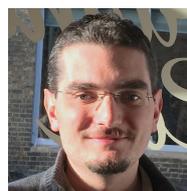

**Andrea Vedaldi** is an associate professor in the Visual Geometry Group at the University of Oxford.